\documentclass{article} % For LaTeX2e
\usepackage{iclr2015,times}
\usepackage[hidelinks]{hyperref}
\usepackage{booktabs}
\usepackage{url}
\usepackage{amsmath}
\usepackage{graphicx}
\usepackage{multirow}
\usepackage{xcolor,colortbl}

\title{
    Leveraging Monolingual Data
        for Crosslingual Compositional Word Representations
}

\author{
    Hubert Soyer \\
    National Institute of Informatics, Tokyo, Japan \\
    {\tt \href{mailto:soyer@nii.ac.jp}{soyer@nii.ac.jp}} \\
    \And
    Pontus Stenetorp\thanks{
        Currently at the University College London.
    } \\
    University of Tokyo, Tokyo, Japan \\
    {\tt \href{mailto:pontus@stenetorp.se}{pontus@stenetorp.se}} \\
    \And
    Akiko Aizawa \\
    National Institute of Informatics, Tokyo, Japan \\
    {\tt \href{mailto:aizawa@nii.ac.jp}{aizawa@nii.ac.jp}}
}

\newcommand{\modelname}{Binclusion }

\definecolor{ColorRed}{rgb}{1,0.5,0.5}
\definecolor{ColorBlue}{rgb}{0.5,0.5,1}

\graphicspath{{./}}

\iclrfinalcopy % Uncomment for camera-ready version

\iclrconference % Uncomment if submitted as conference paper instead of workshop

\begin{document}

\maketitle

\begin{abstract}
In this work, we present a novel neural network based architecture for
inducing compositional crosslingual word representations.
Unlike previously proposed methods, our method fulfills the following three
criteria; it constrains the word-level representations to be compositional,
it is capable of leveraging both bilingual and monolingual data, and it is
scalable to large vocabularies and large quantities of data.
The key component of our approach is what we refer to as a monolingual inclusion
criterion, that exploits the observation that phrases are more closely
semantically related to their sub-phrases than to other randomly sampled
phrases.
We evaluate our method on a well-established crosslingual document
classification task and achieve results that are either comparable, or greatly
improve upon previous state-of-the-art methods.
Concretely, our method reaches a level of $92.7\%$ and $84.4\%$ accuracy for
the English to German and German to English sub-tasks respectively.
The former advances the state of the art by $0.9\%$ points of accuracy,
the latter is an absolute improvement upon the previous state of the art
by $7.7\%$ points of accuracy and an improvement of $33.0\%$ in error reduction.
\end{abstract}

\section{Introduction}

Dense vector representations (embeddings) of words and phrases, as opposed to
discrete feature templates, have recently allowed for notable advances in the
state of the art of Natural Language Processing (NLP)
\citep{socher2013recursive,baroni2014dont}.
These representations are typically induced from large unannotated corpora by
predicting a word given its context \citep{collobert2008unified}.
Unlike discrete feature templates, these representations allow supervised
methods to readily make use of unlabeled data, effectively making
them semi-supervised \citep{turian2010word}.

A recent focus has been on crosslingual, rather than monolingual,
representations.
Crosslingual representations are induced to represent words, phrases,
or documents for more than one language, where the representations are
constrained to preserve representational similarity or can be transformed
between languages
\citep{klementiev2012inducing,mikolov2013mt,hermann2014multilingual}.
In particular, crosslingual representations can be helpful for tasks such as
translation or to leverage training data in a source language when little
or no training data is available for a target language.
Examples of such transfer learning tasks are crosslingual sentiment
analysis \citep{wan2009crosssentiment}
and crosslingual document classification \citep{klementiev2012inducing}.

\citet{mikolov2013mt} induced language-specific word representations,
learned a linear mapping between the language-specific representations using
bilingual word pairs and evaluated their approach for single word translation.
\citet{klementiev2012inducing} used automatically aligned sentences and words
to constrain word representations across languages based on the number of
times a given word in one language was aligned to a word in another language.
They also introduced a dataset for crosslingual document classification and
evaluated their work on this task.
\citet{hermann2014multilingual} introduced a method to
induce compositional crosslingual word representations from sentence-aligned
bilingual corpora. 
Their method is trained to distinguish the sentence pairs given in a
bilingual corpus from randomly generated pairs. 
The model represents sentences as a function of their word representations,
encouraging the word representations to be compositional.
Another approach has been to use auto-encoders and bag of
words representations of sentences that can easily be applied to jointly
leverage both bilingual and monolingual data \citep{chandar2014autoencoder}.
Most recently, \citet{gouws2014bilbowa} extended the Skip-Gram model of
\citet{mikolov2013efficient} to be applicable to bilingual data.
Just like the Skip-Gram model they predict a word in its context,
but constrain the linear combinations of word representations from aligned
sentences to be similar.

However, these previous methods all suffer from one or more of three
short-comings.
\citet{klementiev2012inducing,mikolov2013mt,gouws2014bilbowa} all learn their
representations using a word-level monolingual objective.
This effectively means that compositionality is not encouraged by
the monolingual objective, which may be problematic when composing
word representations for a phrase or document-level task.
While the method of \citet{hermann2014multilingual} allows for arbitrary
composition functions, they are limited to using sentence-aligned bilingual
data and it is not immediately obvious how their method can be extended to
make use of monolingual data.
Lastly, while the method of \citet{chandar2014autoencoder} suffers from
neither of the above issues, their method represents each sentence as
a bag of words vector with the size of the whole vocabulary.
This leads to computational scaling issues and necessitates a vocabulary
cut-off which may hamper performance for compounding languages such as German.

The question that we pose is thus, can a single method
\begin{enumerate}
    \item{Constrain the word-level representations to be compositional.}
    \item{Leverage both monolingual and bilingual data.}
    \item{Scale to large vocabulary sizes
        without greatly impacting training time.}
\end{enumerate}
In this work, we propose a neural network based architecture for creating
crosslingual compositional word representations.
The method is agnostic to the choice of composition function and combines
a bilingual training objective with a novel way of training monolingual
word representations.
This enables us to draw from a plethora of unlabeled monolingual data,
while our method is efficient enough to be trained using roughly
seven million sentences in about six hours on a single-core desktop computer.
We evaluate our method on a well-established document classification task
and achieve results for both sub-tasks that are either comparable or
greatly improve upon the previous state of the art.
For the German to English sub-task our method achieves 84.4\% in accuracy,
an error reduction of 33.0\% in comparison to the previous state of the art.

\section{Model}

\subsection{Inducing Crosslingual Word Representations}

For any task involving crosslingual word representations we
distinguish between two kinds of errors
\begin{enumerate}
    \item{\textbf{Transfer errors} occur due
        to transferring representations between languages.
        Ideally, expressions of the same meaning
        (words, phrases, or documents) should be represented by the same vectors,
        regardless of the language they are expressed in.
        The more different these representations are
        from language 1 ($l_1$) to language 2 ($l_2$),
        the larger the \emph{transfer error}.}
    \item{\textbf{Monolingual errors} occur because the word,
        phrase or document representations within the same language are not
        expressive enough. For example, in the case of classification this
        would mean that the representations do not possess enough
        discriminative power for a classifier to achieve high accuracy.}
\end{enumerate}
The way to attain high performance for any task that involves crosslingual
word representations is to keep both \emph{transfer errors} and
\emph{monolingual errors} to a minimum using representations that are
both expressive and constrained crosslingually.

\subsection{Creating Representations for Phrases and Documents}

Following the work of
\citet{klementiev2012inducing,hermann2014multilingual,gouws2014bilbowa}
we represent each word as a vector and use separate word representations for
each language.
Like \citet{hermann2014multilingual}, we look up the vector representations
for all words of a given sentence in the corresponding lookup table and apply 
a composition function to transform these word vectors into 
a sentence representation.
To create document representations, we apply the same composition function
again, this time to transform the representations of all sentences in a
document to a document representation.
For the majority of this work we will make use of the \emph{addition}
composition function, which can be written as the sum of all word 
representations $w_i$ in a given phrase
\begin{align}
    a([w_1,w_2,\cdots,w_l]) = \sum_{i=1}^{l} w_i
\end{align}

To give an example of another possible candidate composition function,
we also use the bigram based addition (\emph{Bi}) composition function,
formalized as
\begin{align}
    b([w_1,w_2,\cdots,w_l]) = \sum_{i=2}^{l} tanh(w_{i-1},w_i)
\end{align}
where the hyperbolic tangent ($tanh$) is wrapped around every word bigram
to produce intermediate results that are then summed up.
By introducing a non-linear function the \emph{Bi} composition
is no longer a bag-of-vectors function and takes word order into account.

Given that neither of the above composition functions involve any additional
parameters, the only parameters of our model are in fact
the word representations that are shared globally across all training samples.

\subsection{Objective}

Following \citet{klementiev2012inducing} we split our objective into two
sub-objectives, a bilingual objective minimizing the \emph{transfer errors} and
a monolingual objective minimizing the \emph{monolingual errors} for $l_1$ and
$l_2$. We formalize the loss over the whole training set as
\begin{align}
    \label{math:objective}
    L_{total} = \sum_{i=1}^{N_{bi}} L_{bi}(v_i^{l_1}, v_i^{l_2}) +
    \sum_{i=1}^{N_{mono1}} L_{mono} (x_i^{l_1}) +
    \sum_{i=1}^{N_{mono2}} L_{mono} (y_i^{l_2}) +
    \lambda \| \theta \|^2
\end{align}
where $L_{bi}$ is the bilingual loss for two aligned sentences, $v_i$ is a 
sample from the set of $N_{bi}$ aligned sentences in language 1 and 2,
$L_{mono}$ is the monolingual loss which we sum over $N_{mono1}$ sentences
$x_i^{l_1}$ from corpora in language 1 and $N_{mono2}$ sentences $y_i^{l_2}$
from corpora in language 2. We learn the parameters $\theta$, which 
represent the whole set of word representations for both $l_1$ and $l_2$.
The parameters are used in a shared fashion to construct sentence 
representations for both the monolingual corpora and the parts of the 
bilingual corpus corresponding to each language.
We regularize $\theta$ using the squared euclidean norm and scale the
contribution of the regularizer by $\lambda$.

Both objectives operate on vectors that represent composed versions
of phrases and are agnostic to how a phrase is transformed into a vector.
The objective can therefore be used with arbitrary composition functions.

\begin{figure}[t]
    \begin{center}
        \includegraphics[width=0.8\textwidth]{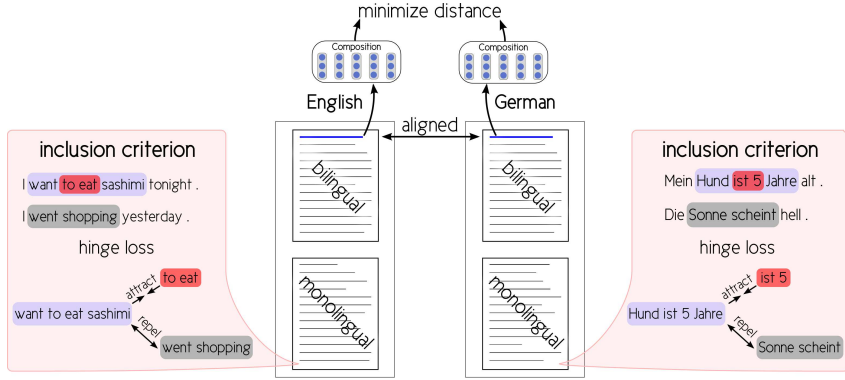}
    \end{center}
    \caption{An illustration of our method.}
    \label{fig:all-illustration}
\end{figure}

An illustration of our proposed method can be found in 
Figure~\ref{fig:all-illustration}.

\subsubsection{Bilingual Objective}

Given a pair of aligned sentences, $s_1^{l_1}$ in $l_1$
and $s_1^{l_2}$ in $l_2$, we first compute their vector representations
$v_1^{l_1}$  and $v_1^{l_2}$ using the composition function.
Since the sentences are either translations of each other or at least
very close in meaning, we require their vector representations to be similar
and express this as minimizing the squared euclidean
distance between $v_1^{l_1}$ and $v_1^{l_2}$.
More formally, we write
\begin{align}
    L_{bi}(v^{l_1}, v^{l_2}) = \| v^{l_1} - v^{l_2} \|^2
\end{align}
for any two vector representations $v^{l_1}$ and $v^{l_2}$ corresponding to
the sentences of an aligned translation pair.

The bilingual objective on its own is degenerate, since setting the vector
representations of all sentences to the same value poses a trivial solution.
We therefore combine this bilingual objective with a monolingual objective.

\subsubsection{Monolingual Objective}

The choice of the monolingual objective greatly influences the
generality of models for crosslingual word representations.
\citet{klementiev2012inducing} use a neural language model to
leverage monolingual data.
However, this does not explicitly encourage compositionality of the 
word representations.
\citet{hermann2014multilingual} achieve good results with a noise-contrastive
objective, discriminating aligned translation pairs from randomly sampled pairs.
However, their approach can only be trained using sentence aligned data,
which makes it difficult to extend to leverage unannotated monolingual data.
\citet{gouws2014bilbowa} introduced BilBOWA combining a bilingual
objective with the Skip-Gram model proposed by \citet{mikolov2013efficient}
which predicts the context of a word given the word itself.
They achieve high accuracy on the $German\rightarrow English$ sub-task of
the crosslingual document classification task introduced by
\citet{klementiev2012inducing}.
\citet{chandar2014autoencoder} presented a
bag-of-words auto-encoder model which is the current state of the art
for the $English\rightarrow German$
sub-task for the same task.
Both the auto-encoder based model and BilBOWA require a sentence-aligned
bilingual corpus, but in addition are capable of leveraging
monolingual data.
However, due to their bag-of-words based nature, their architectures implicitly
restrict how sentence representations are composed from word representations.

We extend the idea of the noise-contrastive objective given by
\citet{hermann2014multilingual} to the monolingual setting
and propose a framework that, like theirs, is agnostic to the choice of
composition function and operates on the phrase level.
However, our framework, unlike theirs, is able to leverage monolingual data.
Our key novel idea is based on the observation that phrases are typically more
similar to their sub-phrases than to randomly sampled phrases.
We leverage this insight using the hinge loss as follows
\begin{align}
    L_{mono}(a) &= L_{mono}(a^{outer}, a^{inner}, b^{noise}) \\
    L_{mono}(a^{outer},a^{inner},b^{noise}) &=
    [\underbrace{max(0, m + \| a^{outer}_{c} - a^{inner}_{c} \|^2 - \| a^{outer}_{c} - b^{noise}_{c} \|^2)}_{\text{hinge loss}} \\
    &\phantom{{}=1} + \| a^{outer}_{c} - a^{inner}_{c} \|^2] \cdot \frac{len(a^{inner})}{len(a^{outer})} \notag
\end{align}
where $m$ is a margin, $a^{outer}$ is a phrase sampled from a sentence,
$a^{inner}$ is a sub-phrase of $a^{outer}$ and $b^{noise}$ is
a phrase extracted from a sentence that was sampled uniformly from
the corpus. 
The start and end positions of both phrases and 
the sub-phrase were chosen uniformly at random within their context and 
constrained to guarantee a minimum length of 3 words.
Subscript $c$ denotes that a phrase has been transformed
into its vector representation.
We add $\| a^{outer}_{c} - a^{inner}_{c} \|^2$ to the hinge loss to reduce the
influence of the margin as a hyperparameter and to make sure that the
we retain an error signal even after the hinge loss objective is satisfied.
To compensate for differences in phrase and sub-phrase length we scale the
error by the ratio between the number of words in the outer phrase and the
inner phrase.
Minimizing this objective captures the intuition stated above; a phrase
should generally be closer to its sub-phrases,
than to randomly sampled phrases.

\begin{figure}[t]
    \begin{center}
        \includegraphics[width=0.75\textwidth]{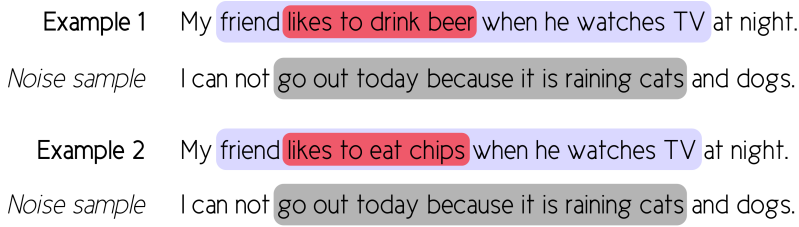}
    \end{center}
    \caption{Examples illustrating the inclusion criterion which we use to leverage
    monolingual text.}
    \label{fig:inclusion-examples}
\end{figure}

The examples in Figure~\ref{fig:inclusion-examples} seek to further clarify
this observation.
In both examples, the blue area represents the outer phrase ($a^{outer}$),
the red area covers the inner sub-phrase ($a^{inner}$), and the gray area
marks a randomly selected phrase in a randomly sampled noise sentence.
The inner workings of the monolingual \emph{inclusion objective} only become
clear when more than one example is considered.
In Example~1, $a^{inner}$ is embedded in the same context as in
Example~2, while in both examples $a^{outer}$ is contrasted with the same noise
phrase.
Minimizing the objective brings the representations of both
\emph{likes to drink beer} and \emph{likes to eat chips} closer to the phrase
they are embedded in and makes them less similar to the same noise sentence.
Since in both examples the outer phrases are very similar, this causes
\emph{likes to drink beer} and \emph{likes to eat chips} to be
similar.
While we picked idealized sentences for demonstration purposes, this relative
notion still holds in practice to varying degrees depending on the choice
of sentences.

In contrast to many recently introduced log-linear models, like the Skip-Gram
model, where word vectors are similar if they appear as the center of similar
word windows, our proposed objective, using addition for composition, encourages
word vectors to be similar if they tend to be embedded in similar phrases.
The major difference between these two formulations manifests itself for words
that appear close or next to each other very frequently.
These word pairs are not usually the center of the same word windows,
but they are embedded together in the same phrases.

For example: the two word central context of ``eat'' is ``to'' and
``chips'', whereas the context of ``chips'' would be ``eat'' and ``when''.
Using the Skip-Gram model this would cause ``chips'' and ``eat'' to be
less similar, with ``chips'' probably being similar to other words related to
food and ``eat'' being similar to other verbs.
Employing the inclusion objective, the representations for ``eat'' and
``chips'' will end up close to each other since they tend to be embedded in the
same phrases.
This causes the word representations induced by the inclusion criterion to be
more topical in nature.
We hypothesize that this property is particularly useful for document
classification.

\section{Experiments}

\subsection{Crosslingual Document Classification}

Crosslingual document classification constitutes a task where a classifier is
trained to classify documents in one language ($l_1$) and is later applied to
documents in a different language ($l_2$).
This requires either transforming the classifier itself to fit the new language
or transforming/sharing representations of the text for both languages.
The crosslingual word and document representations induced using the approach
proposed in this work present an intuitive way to tackle crosslingual
document classification.

Like previous work, we evaluate our method on the crosslingual
document classification task introduced by \citet{klementiev2012inducing}.
The goal is to correctly classify news articles taken from
the English and German sections of the RCV1 and RCV2 corpus
\citep{lewis2004rcv1} into one of four categories:
Economics, Government/Social, Markets, or Corporate.
Maintaining the original setup, we train an averaged perceptron
\citep{collins2002discriminative} for  10 iterations on representations of
documents in one language (English/German) and evaluate its performance on
representations of documents in the corresponding other language
(German/English).

We use the original data and the original implementation of the 
averaged perceptron used by \citet{klementiev2012inducing} to evaluate
the document representations created by our method.
There are different versions of the training set of varying sizes, ranging
from 100 to 10,000 documents, and the test sets for both languages contain
5,000 documents.
Most related work only reports results using the 1,000 documents sized
training set.
Following previous work, we tune the hyperparameters of our model on held out 
documents in the same language that the model was trained on.

\subsection{Inducing Crosslingual Word Representations}

To induce representations using the method proposed in this work, we require at least
a bilingual corpus of aligned sentences.
In addition, our model allows the representations to draw upon monolingual data
from either or both languages.
Like \citet{klementiev2012inducing} we choose EuroParl v7
\citep{koehn2005europarl} as our bilingual corpus and leverage the
English and German parts of the RCV1 and RCV2 corpora as
monolingual resources.
To avoid a testing bias, we exclude all documents that are part of the
crosslingual classification task.
We detect sentence boundaries using pre-trained models of the Punkt tokenizer
\citep{kiss2006unsupervised} shipped with 
NLTK\footnote{\url{http://www.nltk.org/}} and perform tokenization and 
lowercasing with the scripts deployed with the 
cdec decoder\footnote{\url{http://www.cdec-decoder.org/}}.
Following \citet{turian2010word} we remove all English sentences
(and their German correspondences in EuroParl) that have a
$\frac{lowercase}{nonlowercase}$ ratio of less than $0.9$.
This affects mainly headlines and reports with numbers.
In total it reduces the number of sentences in EuroParl by about $255,000$
and the English part of the Reuters corpus by about $8$ million.
Since German features more upper case characters
than English we set the cutoff ratio to $0.7$, which reduces the number
of sentences by around $620,000$.
Further, we replace words that occur less than a certain threshold by an
UNK token.
Corpus statistics and thresholds are reported in Table~\ref{tab:corpora}.
\begin{table}[t]
\caption{Statistics from the corpora used to induce crosslingual
word representations listing type, frequency threshold for turning
tokens into UNKs, the number of sentences, the number of tokens and
the vocabulary size.
The statistics were calculated on the preprocessed versions of the corpora.
}
\label{tab:corpora}
\begin{center}
\begin{tabular}{llllll}
\toprule
                & Type        & UNK threshold & \#sentences & \#tokens & $\vert V\vert$\\
\midrule
EuroParl (EN)   & bilingual   & 2             & 1.66 million & 46 million  & 51,000 \\
EuroParl (DE)   & bilingual   & 2             & 1.66 million & 48 million  & 163,000 \\
Reuters (EN)    & monolingual & 5             & 4.5 million  & 120 million & 114,000\\
Reuters (DE)    & monolingual & 3             & 0.9 million  & 18 million  & 117,000\\
\bottomrule
\end{tabular}
\end{center}
\end{table}

We initialize all word representations with noise samples from a Gaussian
with $\mu = 0, \sigma = 0.1$ and optimize them in a stochastic setting
to minimize the objective defined in Equation \ref{math:objective}.
To speed up the convergence of training we use AdaGrad
\citep{duchi2011adaptive}. We tuned all hyperparameters of our model
and explored learning rates around $0.2$, mini-batch sizes around 40,000,
hinge loss margins around 40 (since our vector dimensionality is 40) and
$\lambda$ (regularization) around $1.0$. We trained all versions
that use the full monolingual data for 25 iterations
(= $25 \times 4.5$ million samples) and the versions only involving bilingual
data for 100 iterations on their training sets.
Training our model\footnote{Our implementation is available at \url{https://github.com/ogh/binclusion}}, implemented in a high-level, dynamic programming language
\citep{bezanson2012julia}, for the largest set of data takes roughly
six hours on a single-core desktop computer.
This can be compared to for example \citet{chandar2014autoencoder} which train
their auto-encoder model for 3.5 days.

\section{Results}

\subsection{Crosslingual Document Classification}

We compare our method to various architectures introduced in previous work.
As these methods differ in their ability to handle monolingual data,
we evaluate several versions of our model using different data sources and
sizes for training.
Also, we follow the lines of previous work and use 40-dimensional
word representations.
We report results when
using
the first 500,000 sentence pairs of EuroParl (\emph{Euro500k}),
the full EuroParl corpus (\emph{EuroFull}),
the first 500,000 sentence pairs of EuroParl and the German and English text
    from the Reuters corpus as monolingual data (\emph{Euro500kReuters}),
and one version using the full EuroParl
    and Reuters corpus (\emph{EuroFullReuters}).
Table~\ref{tab:classification-results} shows results for all these
configurations.
The result table includes previous work as well as the
\emph{Glossed}, the machine translation and the majority class
baselines from \citet{klementiev2012inducing}.

\begin{table}[t]
    \caption{Results for our proposed models, baselines, and related work.
        All results are reported for a training set size of 1,000 documents
    for each language. We refer to our proposed method as \emph{Binclusion}.}
    \label{tab:classification-results}
    \begin{center}
        \begin{tabular}{p{17em}lcc}
    \toprule
    Method  & Training Data & $EN\rightarrow DE$ & $DE\rightarrow EN$ \\
    \midrule
    Machine Translation                                     && 68.1          & 67.4   \\
    Glossed                                                 && 65.1          & 68.6   \\
    Majority Class                                          && 46.8          & 46.8   \\
    \midrule
    I-Matrix\hfill\citep{klementiev2012inducing} & EuroFullReuters  & 77.6          & 71.1   \\
    ADD\hfill\citep{hermann2014multilingual} & Euro500k          & 83.7          & 71.4   \\
    BAE-cr\hfill\citep{chandar2014autoencoder} & Euro500k        & 86.1          & 68.8   \\
    BAE-cr\hfill\citep{chandar2014autoencoder} & Euro500kReuters & 87.9          & 76.7   \\
    BAE-cr\hfill\citep{chandar2014autoencoder} & EuroFullReuters & 91.8 & 74.2   \\
    BilBOWA\hfill\citep{gouws2014bilbowa} & Euro500k             & 86.5          & 75.0   \\
    \midrule
    
    \modelname  &Euro500k                                  & 86.8          & 76.7   \\ 
    \modelname  &EuroFull                                  & 87.8          & 75.7   \\ 
    \modelname  &Euro500kReuters                           & \textbf{92.7}          & \textbf{84.4}   \\ 
    \modelname  (reduced vocabulary) &Euro500kReuters                              & 92.6          & 82.8   \\   
    \modelname  &EuroFullReuters                           & 90.8          & 79.5   \\ 
    \modelname  (Bi) &EuroFullReuters                       & 89.8          & 80.1   \\ 
    \bottomrule
    \end{tabular}
    \end{center}
\end{table}

Our method achieves results that are comparable or improve upon
the previous state of the art for all dataset configurations.
It advances the state of the art for the $EN\rightarrow DE$ 
sub-task by $0.9\%$ points of accuracy and greatly outperforms the 
previous state of the art for the $DE \rightarrow EN$ sub-task,
where it yields an absolute improvement of $7.7\%$ 
points of accuracy.
The latter corresponds to an error reduction of $33.0\%$ in comparison 
to the previous state of the art.

An important observation is that including monolingual data is strongly
beneficial for the classification accuracy.
We found increases in performance to $80.6\%$ for $DE\rightarrow EN$ and
$88.6\%$ accuracy for $EN\rightarrow DE$, even when using as little
as $5\%$ of the monolingual data.
We hypothesize that the key cause of this effect is domain adaptation.
From this observation it is also worth pointing out that our method is on par
with the previous state of the art for the $DE\rightarrow EN$ sub-task 
using no monolingual training data and would improve upon it
using as little as $5\%$ of the monolingual data.
To show that our method achieves high accuracy even with a reduced vocabulary,
we discard representations for infrequent terms and report results using our
best setup with the same vocabulary size as \citet{klementiev2012inducing}.

\subsection{Interesting Properties of the Induced Crosslingual Word Representations}

For a bilingual word representation model that uses monolingual data,
the most difficult cases to resolve are words appearing in
the monolingual data, but not in the bilingual data.
Since the model does not have any kind of direct signal
regarding what translations these words should correspond to,
their location in the vector space is entirely determined
by how the monolingual objective arranges them.
Therefore, looking specifically at these difficult examples presents a
good way to get an impression of how well the monolingual and
bilingual objective complement each other.

In Table~\ref{tab:unseen-english-german},
we list some of the most frequently occurring words that are present
in the monolingual data but not in the bilingual data.
The nearest neighbors are topically strongly related to their corresponding
queries.
For example, the credit-rating agency Standard \& Poor's
(\emph{s\&p}) is matched to rating-related words, \emph{soybeans} is proximal
to crop and food related terms, \emph{forex} features a list of currency
related terms, and the list for \emph{stockholders},
includes \emph{aktion\"are}, its correct German translation.
This speaks strongly in favor of how our objectives complement each other,
even though these words were only observed in the monolingual
data they relate sensibly across languages.

\begin{table}[t]
    \caption{German nearest neighbors for English words that only appear in the monolingual data.}
    \label{tab:unseen-english-german}
    \begin{center}
    \small
    \begin{tabular}{lllll}
    \toprule
     English                 & soybeans           & forex              & s\&p                  & stockholders             \\
    \midrule
     \multirow{5}{*}{German} & mais               & drachme            & ratings               & aktion\"arsschutz        \\
                             & alkoholherstellung & liquidit\"atsfalle & ratingindustrie       & minderheitenaktion\"are  \\
                             & silomais           & bankenliquidit\"at & ratingbranche         & aktion\"arsrechte        \\
                             & genmais            & abnutzung          & ratingstiftung        & aktion\"are              \\
                             & gluten             & pf\"andung         & kreditratingagenturen & minderheitenaktion\"aren \\
    \bottomrule
    \end{tabular}
    \end{center}
\end{table}

To convey an impression of how the induced representations behave, not
interlingually, but within the same language, we list some examples in
Table~\ref{tab:examples-english}.
The semi-conductor chip maker \emph{intel}, is very close to IT-related
companies like \emph{ibm} or \emph{netscape} and also to
microprocessor-related terms.
For the verb \emph{fly}, the nearest neighbors not only include forms like
\emph{flying}, but also related nouns like \emph{airspace} or \emph{air},
underlining the topical nature of our proposed objective.

\begin{table}[t]
    \caption{English words and their nearest neighbors in the
    induced space, demonstrating the topical nature of the word representations.}
    \label{tab:examples-english}
    \begin{center}
    \small
    \begin{tabular}{lllll}
    \toprule
     Query                      & kabul    & intel           & transport      & fly      \\
    \midrule
     \multirow{5}{*}{Neighbors} & taliban  & pentium         & traffic        & air      \\
                                & talibans & microprocessor  & transporting   & flying   \\
                                & taleban  & ibm             & transports     & flight   \\
                                & masood   & microprocessors & dockworkers    & airspace \\
                                & dostum   & netscape        & transportation & naval    \\
    \bottomrule
    \end{tabular}
    \end{center}
\end{table}

\section{Conclusion and Future Work}

In this work we introduced a method that can induce compositional
crosslingual word representations while scaling to large datasets.
Our novel approach for learning monolingual representations
integrates naturally with our bilingual objective and
allows us to make use of sentence-aligned bilingual corpora as well as
monolingual data.
The method is agnostic to the choice of composition function, enabling more
complex (e.g. preserving word order information) ways to compose phrase
representations from word representations.
For crosslingual document classification \citep{klementiev2012inducing} our
models perform comparably or greatly improve upon previously reported results.

To increase the expressiveness of our method we plan to investigate more complex
composition functions, possibly based on convolution to preserve word order information.
We consider the monolingual inclusion objective worthy of further
research on its own and will evaluate its performance in comparison to
related methods when learning word representations from monolingual data.

\section*{Acknowledgements}

This work was supported by the
Data Centric Science Research Commons Project at the
Research Organization of Information and Systems
and by the
Japan Society for the Promotion of Science KAKENHI Grant Number 13F03041.

\bibliography{iclr2015}
\bibliographystyle{iclr2015}

\end{document}